\documentclass[10pt,conference]{IEEEtran}

\usepackage{cite}
\usepackage{graphicx}
\usepackage{amsmath}
\usepackage{amssymb}
\usepackage{booktabs}
\usepackage{array}
\usepackage{xcolor}
\usepackage[hidelinks]{hyperref}
\usepackage{url}
\usepackage{placeins}

\newcolumntype{L}[1]{>{\raggedright\arraybackslash}p{#1}}

\begin{document}

\title{Where the Cost Falls: A Deployment-Aware\\ Adoption Order for Stability Enhancements\\ to Cycle-Consistent Adversarial Networks}

\author{\IEEEauthorblockN{Rowan Hussein\IEEEauthorrefmark{1} and Mohamed Ouf\IEEEauthorrefmark{2}}
\IEEEauthorblockA{\IEEEauthorrefmark{1}University of Ottawa, Ottawa, Canada\\
\IEEEauthorrefmark{2}Queen's University, Kingston, Canada}}

\maketitle

\begin{abstract}
Teams that adopt cycle-consistent adversarial networks for unpaired image-to-image translation meet the same obstacles: adversarial training oscillates or collapses, cycle consistency preserves coarse layout while finer texture drifts, and a single discriminator judging global realism misses local artifacts. Four enhancements address these failures, and they are usually compared on output quality alone. We show that they also divide sharply by where their cost falls, and that this division, which follows from the architecture and not from any particular run, yields an adoption order for teams under a compute or latency budget. A Wasserstein objective with gradient penalty, a VGG19 perceptual loss on the cycle reconstruction, and multi-scale discriminators change training only, so a team can adopt or drop them without altering what ships. Self-attention alone persists into the deployed generator, with memory growing as the square of the feature-map size, which makes it the one component a resource-constrained team should defer. We integrate all four onto a lightly tuned baseline for horse-to-zebra translation, introduced one at a time on a fixed control and then combined, and for each we give the failure mode it targets and how it integrates. We document the collapse and reconstruction-artifact modes the baseline produced, report what visual inspection of saved samples showed for each variant, and report Fr\'echet Inception Distance and Kernel Inception Distance for the combined model. We specify the protocol still needed, covering the individual variants, perceptual similarity, and downstream segmentation, to rank these enhancements on measured evidence.
\end{abstract}


\section{Introduction}
Unpaired image-to-image translation learns a mapping between two visual domains when no aligned pairs are available. CycleGAN established the dominant recipe for this setting by training two generators and enforcing a cycle-consistency constraint, so that an image translated to the other domain and back recovers the original~\cite{zhu2017cyclegan}. CyCADA extended the same idea to unsupervised domain adaptation, combining pixel-level translation with feature-level alignment so that a model trained on a labeled source domain transfers to an unlabeled target~\cite{hoffman2018cycada}. Both methods are now standard building blocks in applied pipelines that need to bridge a visual gap without paired supervision.

Practitioners who deploy these models nonetheless encounter recurring obstacles. Adversarial training oscillates and sometimes collapses, producing repetitive or blank outputs~\cite{arjovsky2017wgan, salimans2016improved}. Cycle consistency preserves coarse structure but allows finer semantic detail, such as the placement and regularity of texture, to drift. A single discriminator focused on global realism can miss local artifacts. These failures matter because the translated images are often an input to a downstream task, where unstable patterns and lost detail degrade performance.

This paper asks which targeted enhancements are worth adding when an engineer starts from a working cycle-consistent baseline, what each one changes in the output, and what each one costs to keep. We study this on the horse-to-zebra task, a benchmark chosen for its clear domain gap and well-understood texture~\cite{zhu2017cyclegan}. We add four enhancements drawn from the generative modeling literature, each aimed at a specific failure mode, and we inspect the results by eye. The cost analysis is the part of this account that transfers beyond the benchmark: it follows from where each component sits in the training loop and in the deployed model, not from the particular domain pair we ran.

We make the following contributions.
\begin{itemize}
\item We describe how four enhancements, a Wasserstein objective with gradient penalty~\cite{gulrajani2017wgangp}, a VGG-based perceptual loss~\cite{johnson2016perceptual, simonyan2015vgg}, self-attention~\cite{zhang2019sagan}, and multi-scale discriminators~\cite{wang2018pix2pixhd}, integrate onto a cycle-consistent baseline.
\item We separate the four by where their cost falls, distinguishing the components that change training only from the one that persists into the deployed generator, and derive an adoption order for teams under a compute or latency budget.
\item We document the collapse and reconstruction-artifact modes the baseline produced, and report what visual inspection of saved samples showed for each variant and for the combined model.
\item We report Fr\'echet Inception Distance and Kernel Inception Distance for the combined model, and specify the remaining evaluation protocol needed to rank the individual enhancements on measured evidence.
\end{itemize}

We intend the paper as an applied reference for engineers stabilizing unpaired translation inside a larger pipeline. It makes no claim of state-of-the-art quality.

\section{Related Work}
\textbf{Generative adversarial networks.} Generative adversarial networks pose image synthesis as a game between a generator and a discriminator~\cite{goodfellow2014gan}. Later work improved the architecture and conditioning~\cite{radford2016dcgan}, and paired translation with conditional adversarial losses~\cite{isola2017pix2pix}. Our study builds on this line but targets the unpaired setting.

\textbf{Unpaired translation and domain adaptation.} CycleGAN~\cite{zhu2017cyclegan}, DiscoGAN~\cite{kim2017discogan}, and DualGAN~\cite{yi2017dualgan} independently proposed cycle consistency for unpaired translation. UNIT and MUNIT added shared latent spaces and multimodal outputs~\cite{liu2017unit, huang2018munit}, and CUT replaced the cycle with a contrastive objective~\cite{park2020cut}. In domain adaptation, feature alignment methods~\cite{long2015dan, ganin2016dann, tzeng2017adda} and pixel-plus-feature methods such as CyCADA~\cite{hoffman2018cycada} reduce the source-target gap. We take a cycle-consistent adversarial model as our baseline and study enhancements to it. We propose no new framework.

\textbf{Training stability.} The Wasserstein GAN reframed the objective around the earth-mover distance~\cite{arjovsky2017wgan}, and the gradient penalty made it practical by enforcing a Lipschitz constraint on the critic~\cite{gulrajani2017wgangp}. Least-squares~\cite{mao2017lsgan} and spectrally normalized~\cite{miyato2018spectral} objectives pursue the same goal by other means. We adopt the gradient-penalty variant.

\textbf{Semantic and perceptual objectives.} Perceptual losses compare deep features from a pretrained network rather than raw pixels~\cite{johnson2016perceptual}, using representations from VGG~\cite{simonyan2015vgg}, and have improved super-resolution~\cite{ledig2017srgan} and served as a learned similarity metric~\cite{zhang2018lpips}. We use a VGG perceptual term to discourage semantic drift.

\textbf{Attention and multi-scale supervision.} Self-attention lets each spatial location attend to the whole feature map~\cite{vaswani2017attention}, and SAGAN introduced it into generative models for long-range coherence~\cite{zhang2019sagan}. Multi-scale discriminators, as in pix2pixHD, judge outputs at several resolutions to balance global structure and local detail~\cite{wang2018pix2pixhd}. We add both self-attention and multi-scale discrimination to the baseline.

\textbf{Evaluation.} Standard quantitative measures for generative models include the Inception Score~\cite{salimans2016improved}, Fr\'echet Inception Distance~\cite{heusel2017fid}, and Kernel Inception Distance~\cite{binkowski2018kid}. We report the latter two for the combined model in Section~\ref{sec:results} and discuss the remaining configurations as the natural next step.

\section{Method: A Baseline and Four Enhancements}
We start from a cycle-consistent adversarial model with two residual-block generators, instance normalization, and PatchGAN discriminators~\cite{zhu2017cyclegan, hoffman2018cycada, isola2017pix2pix, he2016resnet, ulyanov2016instancenorm}. We first apply three preliminary adjustments jointly, described below, and the resulting lightly tuned model is the control for the ablation. We then introduce each of the four enhancements alone on top of this control, so that any change relative to the control can be attributed to that enhancement. The adjustments are bundled rather than ablated individually, so this attribution applies to the four enhancements but not to the adjustments. Table~\ref{tab:enhancements} summarizes the problem each modification targets, its mechanism, and the qualitative effect we observed.

\begin{table*}[t]
\centering
\caption{The preliminary baseline adjustments that define the ablation control (top row) and the four enhancements, the failure mode each targets, its mechanism, and the qualitative effect we observed on horse-to-zebra outputs. The effect column reports visual inspection only.}
\label{tab:enhancements}
\begin{tabular}{L{2.6cm} L{3.6cm} L{4.3cm} L{4.9cm}}
\toprule
\textbf{Enhancement} & \textbf{Failure mode targeted} & \textbf{Mechanism} & \textbf{Observed qualitative effect} \\
\midrule
Baseline adjustments (control) & Unstable updates, limited capacity & Lower generator learning rate, geometric and color augmentation, extra residual block and convolutional layer & Smoother early training and slightly cleaner stripes than the raw baseline \\
\midrule
WGAN-GP~\cite{gulrajani2017wgangp} & Oscillation and mode collapse & Earth-mover objective with a gradient penalty enforcing a Lipschitz critic & Fewer collapsed or blank outputs, more consistent stripe formation \\
VGG perceptual loss~\cite{johnson2016perceptual, simonyan2015vgg} & Loss of fine semantic detail & Match intermediate VGG19 features between each input and its cycle reconstruction & Sharper, better-placed stripes and more distinct object boundaries \\
Self-attention~\cite{zhang2019sagan} & Locally inconsistent patterns & Attention over the full feature map inside the generator & Stripe orientation more coherent across the body \\
Multi-scale discriminators~\cite{wang2018pix2pixhd} & Detail missed by a single scale & Discriminators at coarse and fine resolutions contribute jointly & Reduced small-scale artifacts, crisper coat texture \\
\bottomrule
\end{tabular}
\end{table*}

The control combines three low-cost adjustments a practitioner would try first: the generator learning rate is lowered from $2\times10^{-4}$ to $1\times10^{-4}$ under the Adam optimizer~\cite{kingma2015adam}, the source images receive random rotations, flips, and color jitter, and the generator gains one residual block and a 256-filter convolutional layer with a LeakyReLU activation. On top of this control we add each enhancement alone. WGAN-GP replaces the standard adversarial objective with the Wasserstein distance and a gradient penalty that keeps the critic Lipschitz, with the sigmoid removed and the penalty coefficient set to the recommended value of 10~\cite{arjovsky2017wgan, gulrajani2017wgangp}. The perceptual term matches intermediate VGG19 features between each input and its cycle reconstruction, since the unpaired setting offers no paired target, and complements the pixel-space cycle loss~\cite{johnson2016perceptual, simonyan2015vgg}. Self-attention lets every spatial position attend to all others before decoding, addressing the limited receptive field of convolutions~\cite{zhang2019sagan, vaswani2017attention}. Multi-scale discriminators judge the output at coarse and fine resolutions at once, so the generator must satisfy a critic of global layout and a critic of local texture~\cite{wang2018pix2pixhd}. Table~\ref{tab:enhancements} gives the failure mode, mechanism, and observed effect of each.

The combined model trains with all four enhancements together. In the samples we inspected the components did not visibly interfere, but we do not claim it as a demonstrated property, since they were developed under different conventions and a controlled study would be needed to disentangle their interactions~\cite{zhang2019sagan, miyato2018spectral}.

\subsection{Where each cost falls}
\label{sec:cost}
Each enhancement carries an engineering cost that a team integrating translation into a larger pipeline must weigh, and these costs differ in kind as well as in magnitude. The gradient penalty adds a gradient computation through the critic at every discriminator update, and the perceptual term adds forward passes through a frozen VGG19 network together with its memory footprint. Both modify training only, so a team can adopt them, and abandon them, without changing what ships. Multi-scale discrimination multiplies the number of discriminators to train, which raises training cost with the number of scales but likewise leaves the deployed generator untouched. Self-attention is the exception. It stores weights over all pairs of spatial positions, so its memory cost grows with the square of the feature-map size, and unlike the other three it remains in the deployed generator and is therefore paid at every inference. We did not profile these overheads, but they follow from where each component sits in the training loop and in the shipped model, so they hold beyond the particular runs we performed, and they imply an adoption order. A team under a compute or latency budget should trial the three training-only changes first and treat self-attention as a separate decision, justified against inference cost instead of bundled with the rest.

\section{Experimental Setup}
We ran all experiments on Google Colab Pro using the public horse-to-zebra dataset from the CycleGAN release~\cite{zhu2017cyclegan}. We trained the control and each variant for approximately 100 epochs under identical data and schedule, with each run taking roughly six hours. This is about half of the 200-epoch schedule used in the original CycleGAN work~\cite{zhu2017cyclegan}, so undertraining may contribute to the baseline instability we describe, a confound the qualitative design cannot rule out. Because of the compute budget we studied a single source-target pair and a single seed, which we treat as a limitation. This is not a controlled multi-run comparison. We introduced the enhancements one at a time, retrained from the same starting configuration, and saved sample translations at the end of training. Our analysis combines a visual inspection of these samples with Fr\'echet Inception Distance and Kernel Inception Distance computed for the combined model.

\section{Results}
\label{sec:results}

\subsection{Quantitative evaluation}
Table~\ref{tab:metrics} reports Fr\'echet Inception Distance~\cite{heusel2017fid} and Kernel Inception Distance~\cite{binkowski2018kid} for the combined model on horse-to-zebra translation. We computed these metrics for the combined model only. Without values for the control or for the four single-variable variants, the table fixes the quality of the composed configuration but does not attribute any part of that result to an individual enhancement, and it does not establish an improvement over the control.

\begin{table}[t]
\centering
\caption{Fr\'echet Inception Distance and Kernel Inception Distance for the combined model on horse-to-zebra translation. Lower is better for both. We did not compute these metrics for the control or for the four single-variable variants.}
\label{tab:metrics}
\begin{tabular}{lrr}
\toprule
\textbf{Configuration} & \textbf{FID} $\downarrow$ & \textbf{KID} ($\times 10^{-3}$) $\downarrow$ \\
\midrule
Combined model & 60 & 50 \\
\bottomrule
\end{tabular}
\end{table}

\subsection{Qualitative observations}
\begin{figure}[t]
\centering
\includegraphics[width=\columnwidth]{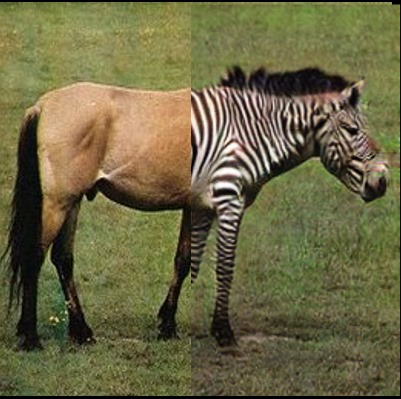}\\
\makebox[0.5\columnwidth]{\footnotesize (a) Input}\makebox[0.5\columnwidth]{\footnotesize (b) Enhanced model}
\caption{Representative horse-to-zebra translation. (a) Input horse. (b) Zebra generated by the enhanced model, with stripes following the body contour and the background retained.}
\label{fig:teaser}
\end{figure}

Figure~\ref{fig:teaser} shows a representative horse-to-zebra translation from the enhanced model, with the input horse on the left and the generated zebra on the right. The generated stripes follow the contour of the body and the head, and the background is preserved. This is the behavior we want from a working translation and it frames the comparisons that follow.

\begin{figure*}[!t]
\centering
\includegraphics[width=0.82\textwidth]{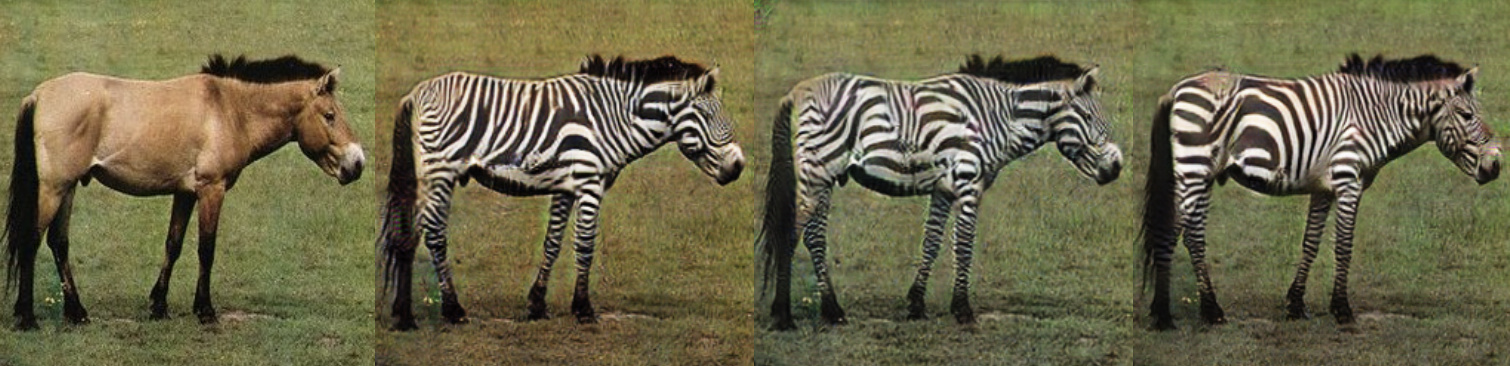}\\
\makebox[0.205\textwidth]{\footnotesize (a) Input}\makebox[0.205\textwidth]{\footnotesize (b) Earlier variant}\makebox[0.205\textwidth]{\footnotesize (c) Intermediate variant}\makebox[0.205\textwidth]{\footnotesize (d) Later variant}
\caption{Input horse (a) and horse-to-zebra outputs (b) to (d) from three model variants at successive stages of the enhancement sequence. Stripe regularity and alignment with the body improve from left to right. The saved samples do not identify the exact configuration behind each output panel, so the figure documents the overall progression only.}
\label{fig:progression}
\end{figure*}

\begin{figure*}[!t]
\centering
\includegraphics[width=0.82\textwidth]{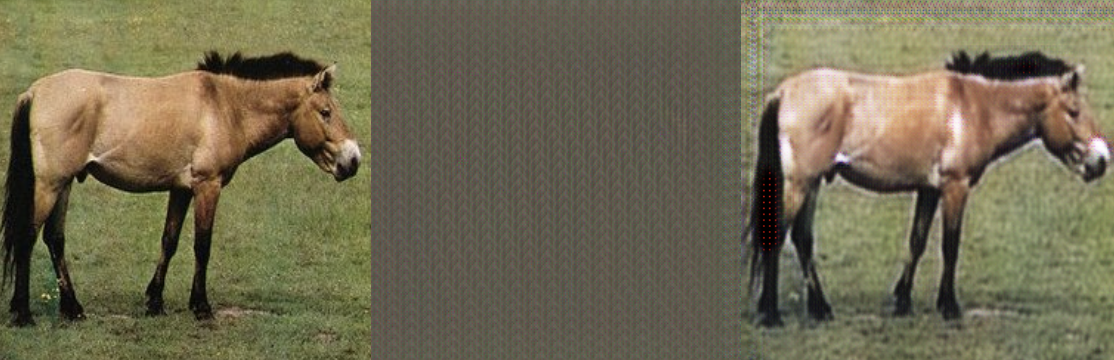}\\
\makebox[0.273\textwidth]{\footnotesize (a) Input}\makebox[0.273\textwidth]{\footnotesize (b) Collapsed translation}\makebox[0.273\textwidth]{\footnotesize (c) Artifacted reconstruction}
\caption{A failure case from the baseline configuration. (a) Input horse, (b) a translation collapsed to a near-uniform texture, and (c) a cycle reconstruction with a faint duplicate and grid-like artifacts. Such collapses motivate the stability enhancements.}
\label{fig:failure3}
\end{figure*}

\begin{figure*}[!t]
\centering
\includegraphics[width=0.82\textwidth]{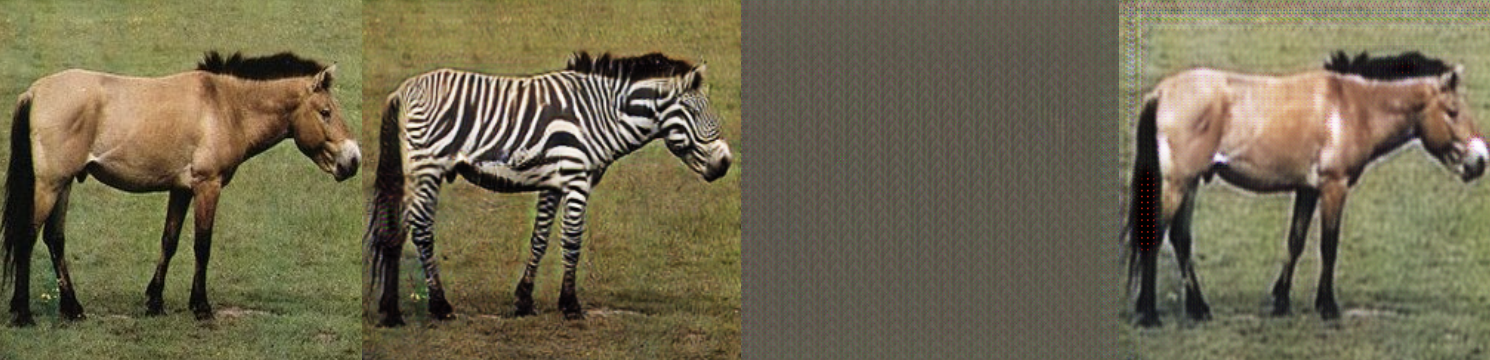}\\
\makebox[0.205\textwidth]{\footnotesize (a) Input}\makebox[0.205\textwidth]{\footnotesize (b) Translation}\makebox[0.205\textwidth]{\footnotesize (c) Collapsed panel}\makebox[0.205\textwidth]{\footnotesize (d) Artifacted reconstruction}
\caption{A second failure case from the baseline configuration. (a) Input horse, (b) a successful translation, (c) a collapsed panel, and (d) a reconstruction with artifacts. Multi-scale discrimination and the Wasserstein objective target the small-scale corruption and instability visible here.}
\label{fig:failure4}
\end{figure*}

\begin{figure*}[!t]
\centering
\includegraphics[width=0.82\textwidth]{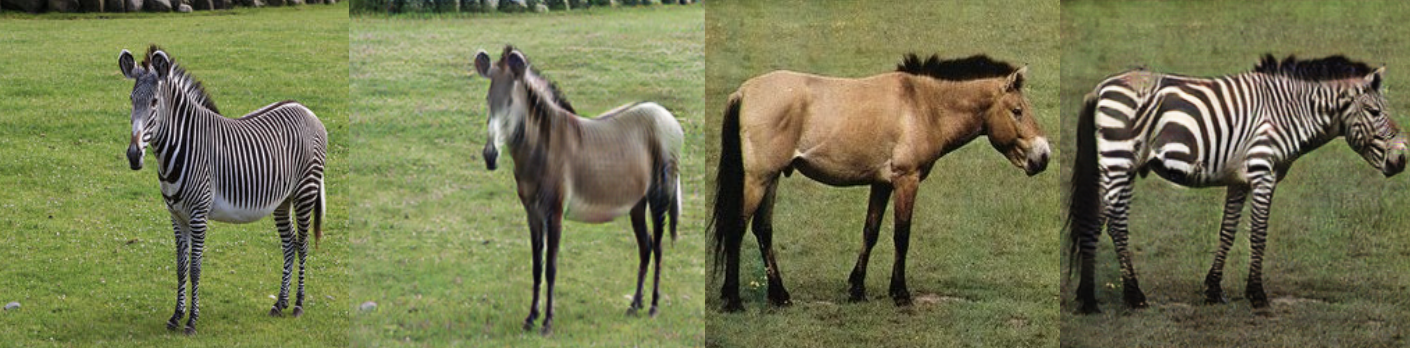}\\
\makebox[0.205\textwidth]{\footnotesize (a) Zebra input}\makebox[0.205\textwidth]{\footnotesize (b) Horse output}\makebox[0.205\textwidth]{\footnotesize (c) Horse input}\makebox[0.205\textwidth]{\footnotesize (d) Zebra output}
\caption{Bidirectional examples. Zebra-to-horse, (a) to (b), removes stripes and yields a plausible but desaturated coat, while horse-to-zebra, (c) to (d), adds strong, well-oriented stripes. The two directions differ in difficulty.}
\label{fig:bidir}
\end{figure*}

Figure~\ref{fig:progression} places an input horse next to horse-to-zebra outputs from three model variants at successive stages of the enhancement sequence. Reading left to right, the stripe pattern becomes more regular and better aligned with the body, and the transition between striped and unstriped regions grows cleaner. The saved samples do not identify the exact configuration behind each panel, so we read the figure as evidence of the overall progression only. The per-enhancement observations in Table~\ref{tab:enhancements} rest on our inspection of samples throughout the study, which went beyond this figure.

Figures~\ref{fig:failure3} and~\ref{fig:failure4} illustrate the failure modes that motivated the stability and detail enhancements, drawn from runs of the baseline configuration. Figure~\ref{fig:failure3} shows an input, a translation that collapsed to a near-uniform texture with no animal, and a cycle reconstruction with a faint duplicate and grid-like artifacts. Figure~\ref{fig:failure4} shows an input, a successful translation, a collapsed panel, and an artifacted reconstruction. These collapsed and corrupted panels are the concrete form of the instability that the Wasserstein objective and multi-scale discriminators are meant to reduce. In our inspection, such collapses appeared less often once those components were in place, but we did not count outcomes systematically and so report no rates.

Figure~\ref{fig:bidir} shows translations in both directions. Horse-to-zebra outputs carry strong, well-oriented stripes, while zebra-to-horse outputs remove the stripes and produce a plausible coat, though with some desaturation and softness. The asymmetry is expected, since adding a regular texture is a different problem from removing one, and it points to per-direction tuning as useful future work.

Across the outputs we reviewed, the combined model produced the most consistent stripe patterns and the fewest collapsed or heavily corrupted outputs. We did not count outcomes across runs, so we report it as an impression from a limited sample.

\section{Discussion}
Each enhancement targets a specific, nameable failure of the cycle-consistent baseline, and in our sample each produced a visible change consistent with its intended effect: fewer collapses with the Wasserstein objective and gradient penalty, sharper and better-placed stripes with the perceptual loss, more globally consistent orientation with self-attention, and crisper texture with multi-scale discriminators. The combined model inherited these behaviors without any single component visibly dominating.

The central limitation is that our observations come from a single seed and a single domain pair, and that we measured only the combined model. With no metrics for the control or for the single-variable variants, the per-enhancement ranking rests on visual inspection, which is subject to selection and confirmation bias and cannot separate a real gain from run-to-run variance. We therefore make no comparative or state-of-the-art claim, and we make no claim about downstream segmentation or detection performance, since we did not measure it. We also report only the configuration details we can verify from our records, omitting the individual loss weights, the batch size and image resolution, and the exact insertion point of the attention module, so this is an integration account short of a complete reproduction package. The same gap applies to the evaluation: we do not specify the number of generated images or the reference split behind the reported distances, so those values locate the combined model within this study and should not be read against published figures for this benchmark. The cost analysis in Section~\ref{sec:cost} is the part least exposed to these gaps, since it turns on where each component sits in the pipeline, independent of the values we used.

The remedy is to complete the quantitative protocol we began. First, compute Fr\'echet Inception Distance~\cite{heusel2017fid} and Kernel Inception Distance~\cite{binkowski2018kid} between generated and real target images for the control and for each single-variable variant, as we did for the combined model, across several seeds and with confidence intervals. Second, use a learned perceptual similarity such as LPIPS~\cite{zhang2018lpips} to quantify detail preservation. Third, run a downstream test that reflects the applied motivation: train a segmentation or detection model on the translated domain and measure its accuracy, following the domain-adaptation evaluation logic of CyCADA~\cite{hoffman2018cycada}. Only such measurements can confirm or overturn the qualitative ranking we observed, and they would let the recipe be tuned instead of assembled by inspection.

Beyond measurement, the asymmetry between the two translation directions in Figure~\ref{fig:bidir} suggests direction-specific loss weighting, and the single domain pair invites replication on the other CycleGAN benchmarks. Stabilized translation also matters wherever labeled target data is scarce and a pipeline must bridge a visual gap before a downstream task, which is the applied setting that motivated this study.

\section{Conclusion}
We reported on integrating four enhancements into a lightly tuned cycle-consistent adversarial baseline for unpaired horse-to-zebra translation. A Wasserstein objective with gradient penalty, a VGG perceptual loss, self-attention, and multi-scale discriminators each target a distinct failure of the baseline, and combining them yielded the most consistent and least corrupted outputs among the samples we inspected. Our claims about output quality are limited to what visual inspection of a small sample supports. The cost finding is more durable: three of the four components change training only and can be adopted or dropped without altering what ships, while self-attention persists into the deployed generator with memory quadratic in feature-map size, which gives a team under budget a defensible order in which to try them. We contribute the composed configuration together with its measured Fr\'echet Inception Distance and Kernel Inception Distance, a description of its failure modes and of where each component's cost falls, and a concrete protocol covering the individual variants, perceptual similarity, and downstream segmentation for future work to test these observations and to tune the configuration on a firmer basis.

\bibliographystyle{IEEEtran}
\bibliography{main}

\end{document}